\documentclass[12pt,a4paper,doc]{apa6}
\usepackage[british,UKenglish,USenglish,american]{babel}

\usepackage[utf8]{inputenc}    
\usepackage[T1]{fontenc}       

\newif\ifinteract 
\interactfalse 

\usepackage{amsmath}
\usepackage{enumerate}
\usepackage{graphicx}
\usepackage{multirow}
\usepackage{nameref}
\usepackage{hyperref}
\usepackage{palatino}
\usepackage{fancyvrb}
\usepackage{color}
\usepackage{xcolor}
\graphicspath{{figs/}}
\usepackage{defs}
\usepackage{url}
\usepackage{makecell} 
\usepackage{float}
\usepackage{csquotes}
\usepackage{placeins}

\ifinteract
\usepackage{setspace} 
\usepackage{lineno} 
\fi

\definecolor{elmeucolor}{RGB}{255,0,0}

\ifinteract
\usepackage[
    backend=biber, 
    style=apa,
    citestyle=apa,
    sorting=nyt,
    sortlocale=en_US,
    sortcites=true,
    url=true,
    doi=false,
    eprint=false,
    safeinputenc
]{biblatex}
\DeclareLanguageMapping{american}{american-apa}
\else
\newcommand{\parencite}[1]{\cite{#1}}
\newcommand{\textcite}[1]{\cite{#1}}
\fi

\title{The Polysemy of the Words that Children Learn over Time}
\shorttitle{The Polysemy of the Words that Children Learn}

\fiveauthors{\large Bernardino Casas  ~ bcasas@cs.upc.edu $^1$}
{\large Neus Catal\`a  ~ ncatala@cs.upc.edu $^2$}
{\large Ramon Ferrer-i-Cancho  ~ rferrericancho@cs.upc.edu $^1$}
{\large Antoni Hern\'andez-Fern\'andez  ~ antonio.hernandez@upc.edu $^{1,3}$}
{\large Jaume Baixeries  ~ jbaixer@cs.upc.edu $^1$}

\fiveaffiliations{~}
{~}
{~}
{~}
{~}

\note{1. Complexity \& Quantitative Linguistics Lab, 
	Departament de Ci\`encies de la Computaci\'o, 
	Laboratory for Relational Algorithmics, 
	Complexity and Learning (LARCA), Universitat Polit\`ecnica de Catalunya, 
	Barcelona, Catalonia.\\
	2. Complexity \& Quantitative Linguistics Lab, 
	Departament de Ci\`encies de la Computaci\'o, 
	Center for Language and Speech Technologies and Applications 
	(TALP Research Center), Universitat Polit\`ecnica de Catalunya, 
	Barcelona, Catalonia.\\
	3. Institut de Ci\`encies de l$'$Educaci\'o, 	
	Universitat Polit\`ecnica de Catalunya, Barcelona, Catalonia.}

\begin{document}

\maketitle
\newpage

\ifinteract
\else

\def\figuraPolysemyAllWordNet{
\begin{figure}[htp!]
	\centering
	\includegraphics[width=\textwidth]{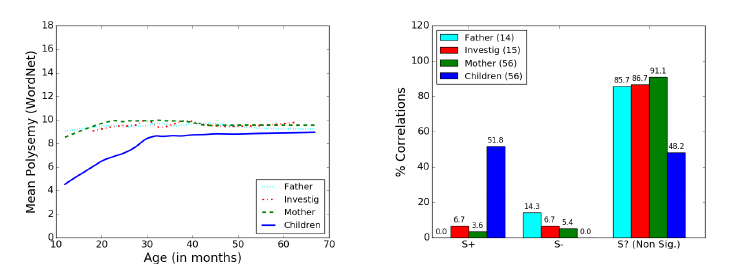}
\caption{Evolution of the \textbf{WordNet polysemy} by children's age in content words.
Left: average dependency between mean polysemy and child age for each role.
The breakpoint in the growth of the mean polysemy of children is located at $31.6 \pm 0.1$ months.
Right: percentage of correlations between mean polysemy and age ($S_+$: positive significant; $S_-$ negative significant; $S_?$: non-significant).
The value above every bar is the percentage of correlations for this role.
The number of conversations used to calculate the correlations for every role appears between parentheses in the legend.
Sample length: 50 tokens. }
\label{fig:eng_edat_sinsets_all_50}
\end{figure}
}

\def\figuraPolysemyAllSemCor{
\begin{figure}[htp!]
	\includegraphics[width=\textwidth]{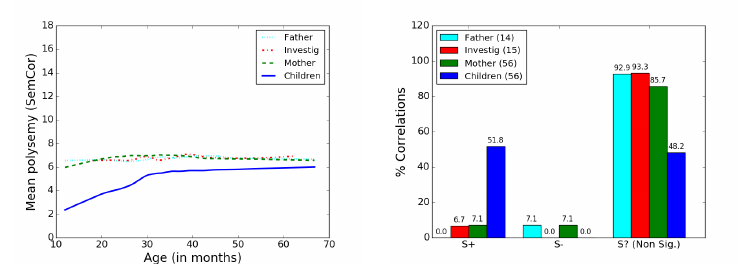}
	\caption{Evolution of the \textbf{SemCor polysemy} by children's age in content words.
	The format and sample length are the same as
	in Figure \ref{fig:eng_edat_sinsets_all_50}.
	The breakpoint in the growth of mean polysemy of children is located at $31.4 \pm 0.1$ months.
	}
	\label{fig:eng_edat_sinsets_semcor_all_50}
\end{figure}
}

\def\figuraBoxPlotAllCorrelations{
	\begin{figure}[htp!]
		\includegraphics[width=\textwidth]{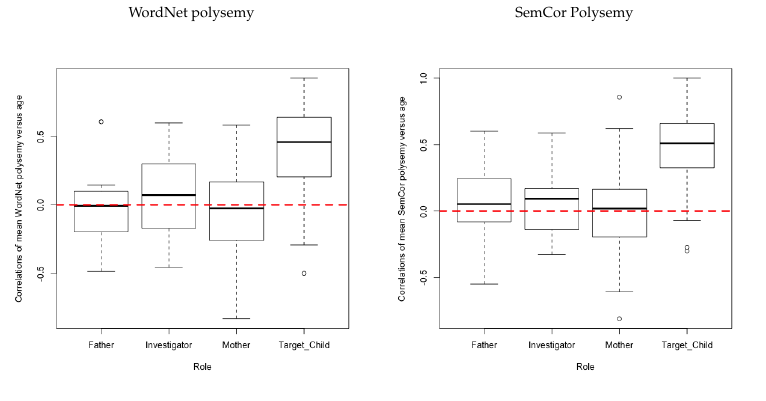}
		\caption{Box plot of the correlation between polysemy and child age (left WordNet polysemy, right SemCor polysemy)
			for all categories (nouns, verbs, adjectives and adverbs). The thick line indicates the median. The top and the bottom of the box indicate the $75\%$ and the $25\%$ percentile, respectively. 
			The dashed red line is a guide to the eye indicating zero correlation.    
		}
		\label{fig:eng_boxplot_corr}
	\end{figure}
}

\def\figuraPolysemyCategoryWordNet{
\begin{figure}[htp!]
	\includegraphics[width=\textwidth]{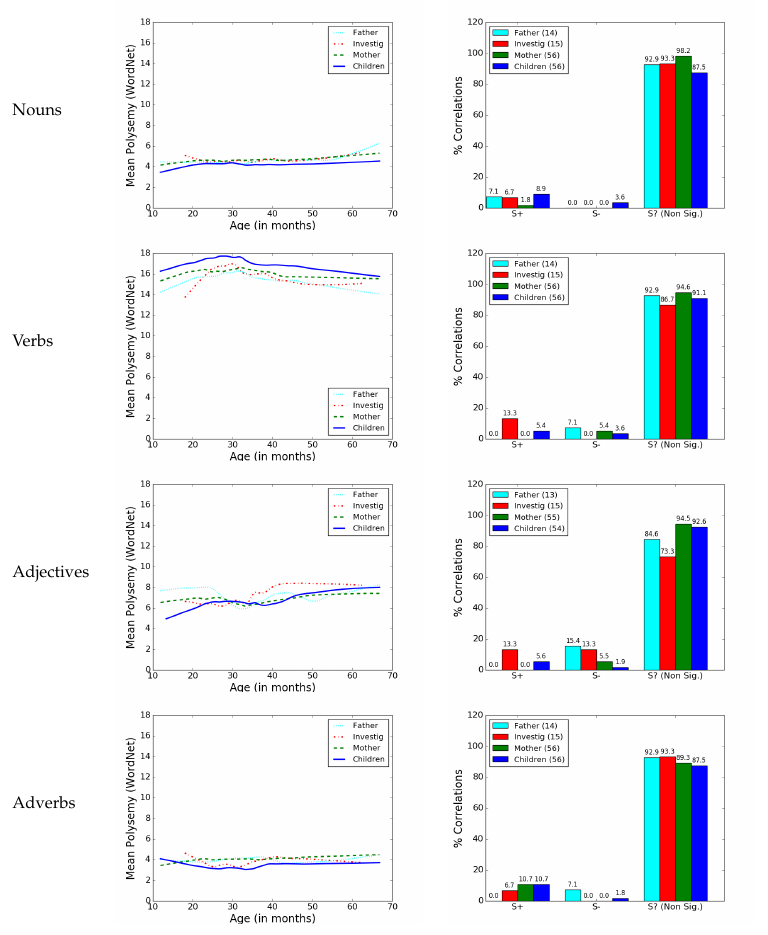}
\caption{Evolution of the \textbf{WordNet polysemy} by child age for only nouns,
only verbs, only adjectives and only adverbs.
Left figures: average dependency between mean polysemy and child age of each role.
Right figures: percentage of correlations between mean polysemy and age
($S_+$: positive significant; $S_-$ negative significant; $S_?$: non-significant).
The value above every bar is the percentage of correlations for this role.
The number of conversations used to calculate the correlations for every role appears between parentheses in the legend.
Sample length: 50 tokens.}
\label{fig:eng_edat_sinsets_other_50}
\end{figure}
}

\def\figuraPolysemyCategorySemCor{
\begin{figure}[htp!]
	\includegraphics[width=\textwidth]{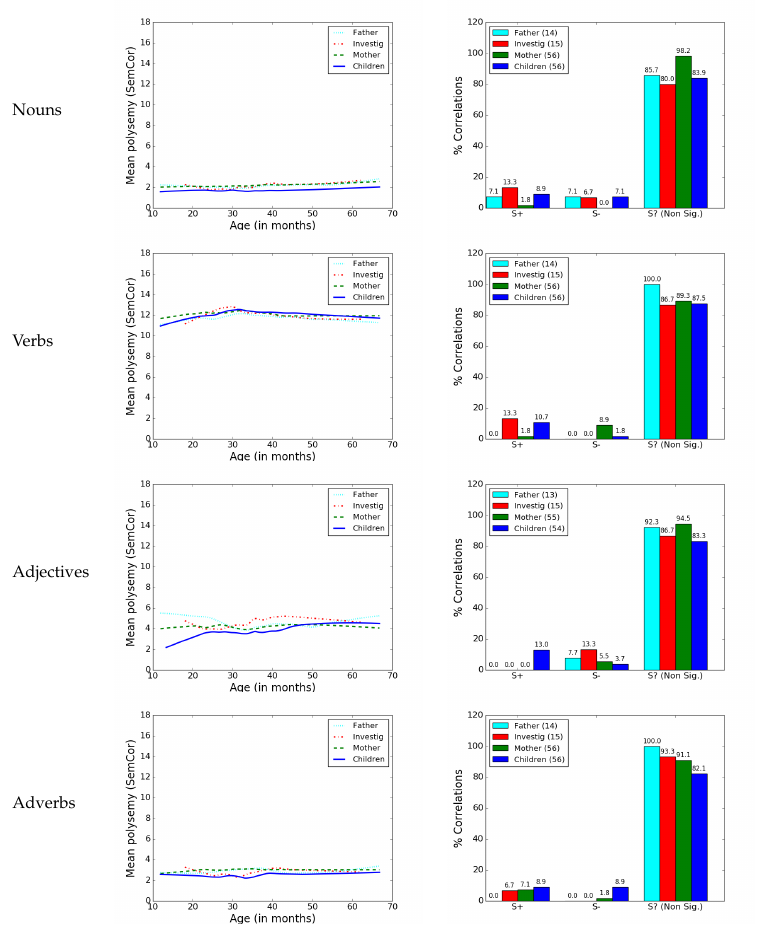}
\caption{Evolution of the \textbf{SemCor polysemy} by child age for only nouns,
			only verbs, only adjectives and only adverbs.
			The format and sample length are the same as
			in Figure \ref{fig:eng_edat_sinsets_other_50}.
		}
\label{fig:eng_edat_sinsets_semcor_other_50}
\end{figure}
}

\def\figuraBoxPlotCategoryCorrelations{
	\newcommand{\amplada}{.4}	
	\begin{figure}[htp!]
		\includegraphics[width=\textwidth]{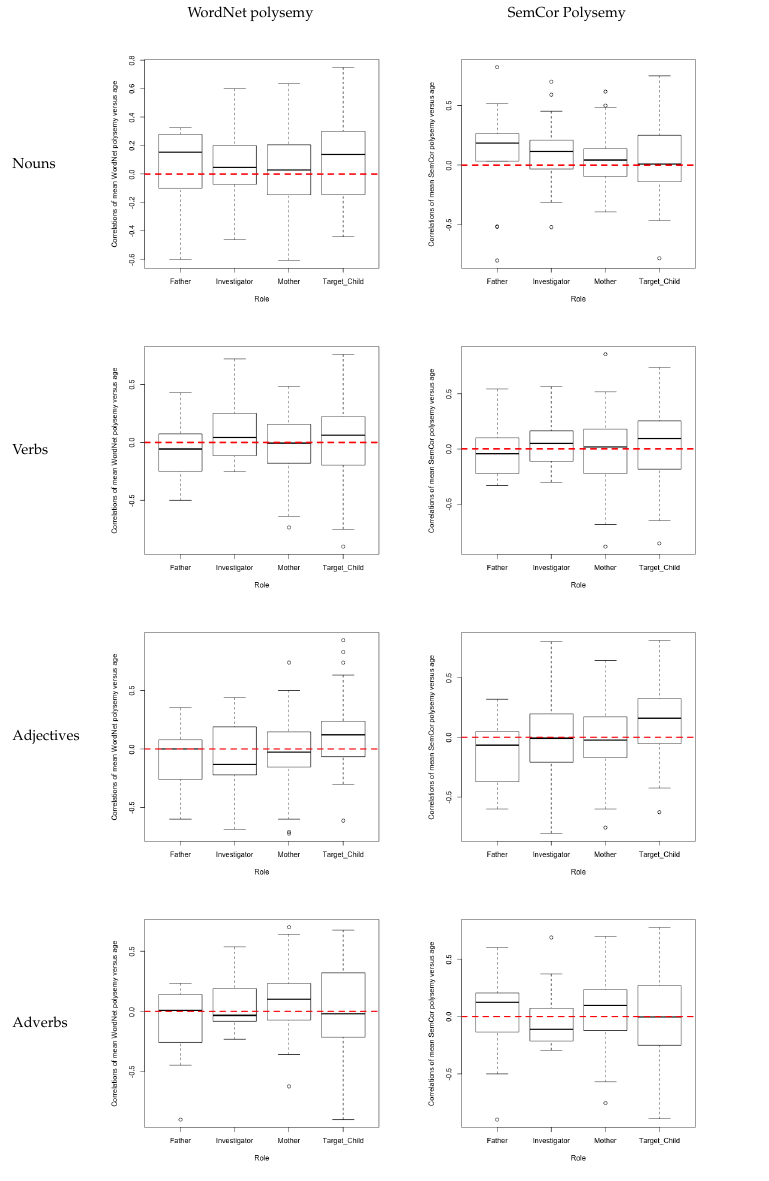}
		\caption{Box plot of the correlation between polysemy and child age (left WordNet polysemy, right SemCor polysemy) for nouns, verbs, adjectives and adverbs separately. The format is the same as that of Fig. \ref{fig:eng_boxplot_corr}.
		}
		\label{fig:eng_boxplot_corr_cat}
	\end{figure}
}

\def\figuraTRES{
\begin{figure}[htp!]
	\includegraphics[width=\textwidth]{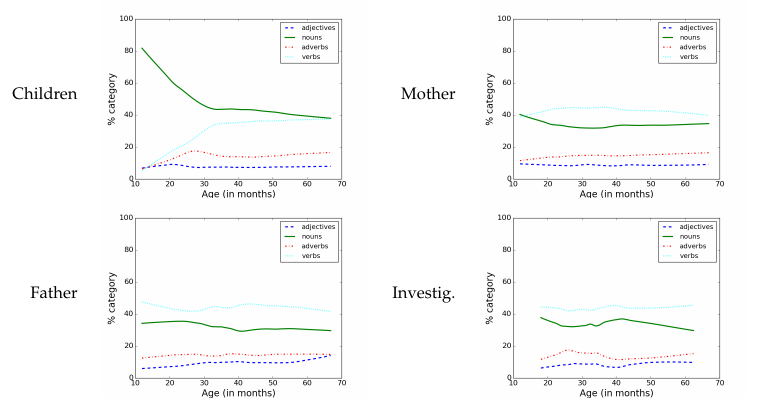}
\caption{Evolution of the percentage of use of the syntactic categories
		(adjectives, nouns, adverbs and verbs) by child age for each role
		(children, mother, father and investigator).
		Sample length: 50 tokens.
		The breakpoint for the percentage of nouns in children is $30.0 \pm 0.1$ months
		and for the percentage of verbs in children is $33.0 \pm 0.1$ months.}
\label{fig:eng_edat_category}
\end{figure}
}

\def\figuraQUATRE{
\begin{figure}[htp!]
	\includegraphics[width=\textwidth]{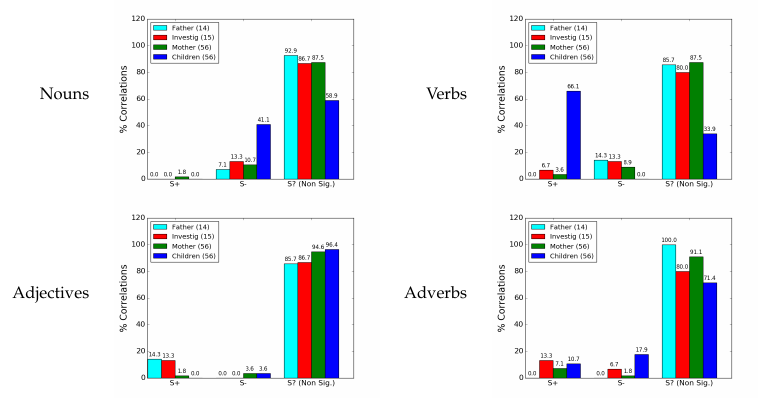}

	\caption{Percentage of correlations
	($S_+$: positive significant; $S_-$ negative significant; $S_?$: non-significant)
	of each syntactic category by child age for each role
	(children, mother, father and investigator).
	The value above every bar is the percentage of correlations for this role.
	The number of conversations used to calculate the correlations for every role appears between parentheses in the legend.
	Sample length: 50 tokens.}
	\label{fig:eng_edat_category_barra}
\end{figure}
}

\fi

\def\taulaRawDataWordNet{
\begin{table}[hp!]
	\begin{center}
\begin{tabular}{|c||c||c||c|c||c|c|c|}
\hline
Category & Role & $N$ & $C_+$ & $C_-$ & $S_+$ & $S_-$ & $S_?$ \\
\hline
\hline
\multirow{4}{*}{All categories}
& Children & 56 & $\uparrow$ 51 & $\downarrow$ 5 & $\uparrow$ 29 &  0 & $\downarrow$ 27  \\
\cline{2-8}
& Mother & 56 &  27 &  29 &  2 &  3 &  51  \\
\cline{2-8}
& Father & 14 &  7 &  7 &  0 &  $\uparrow$ 2 & 12  \\
\cline{2-8}
& Investigator & 15 &  8 &  7 &  1 &  1 &  13  \\
\hline
\hline
\multirow{4}{*}{Nouns}
& Children & 56 & $\uparrow$ 37 & $\downarrow$ 19 & $\uparrow$ 5 &  2 & $\downarrow$ 49  \\
\cline{2-8}
& Mother & 56 &  32 &  24 &  1 &  0 &  55  \\
\cline{2-8}
& Father & 14 &  9 &  5 & 1 &  0 & 13  \\
\cline{2-8}
& Investigator & 15 &  9 &  6 &  1 &  0 &  14  \\
\hline
\hline
\multirow{4}{*}{Verbs}
& Children & 56 & $\uparrow$ 33 & $\downarrow$ 23 &  3 &  2 &  51  \\
\cline{2-8}
& Mother & 56 &  28 &  28 &  0 &  3 &  53  \\
\cline{2-8}
& Father & 14 &  7 &  7 &  0 &  1 &  13  \\
\cline{2-8}
& Investigator & 15 &  9 &  6 & $\uparrow$ 2 &  0 &  13  \\
\hline
\hline
\multirow{4}{*}{Adjectives}
& Children & 54 & $\uparrow$ 38 & $\downarrow$ 16 &  3 &  1 &  50  \\
\cline{2-8}
& Mother & 55 & $\downarrow$ 22 &  $\uparrow$ 33 &  0 & 3 &  52  \\
\cline{2-8}
& Father & 13 &  7 &  6 &  0 & $\uparrow$ 2 &  11  \\
\cline{2-8}
& Investigator & 15 &  6 &  9 & $\uparrow$ 2 & $\uparrow$ 2 & $\downarrow$ 11  \\
\hline
\hline
\multirow{4}{*}{Adverbs}
& Children & 56 &  27 &  29 & $\uparrow$ 6 &  1 & $\downarrow$ 49  \\
\cline{2-8}
& Mother & 56 & $\uparrow$ 37 & $\downarrow$ 19 & $\uparrow$ 6 &  0 & $\downarrow$ 50  \\
\cline{2-8}
& Father & 14 &  7 &  7 &  0 &  1 &  13  \\
\cline{2-8}
& Investigator & 15 &  6 &  9 &  1 &  0 &  14  \\
\hline
\end{tabular}
\caption{Raw data on the evolution of \textbf{WordNet polysemy} over time taking into account
	the syntactic category and role.
	$N$: Number of individuals;
	$C_+$: Positive correlations;
	$C_-$: Negative correlations;
	$S_+$: Positive significant;
	$S_-$: Negative significant;
	$S_{?}$: Neither positive nor negative significant.
	Arrows indicate if the counts are significantly high ($\uparrow$)
	or significantly low ($\downarrow$) according to a binomial test
	for a given category and role (see the Section \nameref{sec:materialsandmethods}
	for further details about this test).
	Sample length: 50 tokens.
}
\label{tab:eng_edat_sinsets_binom_50}
\end{center}
\end{table}
}

\def\taulaRawDataSemCor{
\begin{table}[hp!]
	\begin{center}
\begin{tabular}{|c||c||c||c|c||c|c|c|}
\hline
Category & Role & $N$ & $C_+$ & $C_-$ & $S_+$ & $S_-$ & $S_?$ \\
\hline
\hline
\multirow{4}{*}{All categories}
& Children & 56 & $\uparrow$ 53 & $\downarrow$ 3 & $\uparrow$ 29 &  0 & $\downarrow$ 27  \\
\cline{2-8}
& Mother & 56 &  30 &  26 & $\uparrow$ 4 & $\uparrow$ 4 &  $\downarrow$ 48  \\
\cline{2-8}
& Father & 14 &  9 &  5 &  0 &  1 &  13  \\
\cline{2-8}
& Investigator & 15 &  9 &  6 &  1 &  0 &  14  \\
\hline
\hline
\multirow{4}{*}{Nouns}
& Children & 56 & 29 & 27 & $\uparrow$ 5 & $\uparrow$ 4 & $\downarrow$ 47  \\
\cline{2-8}
& Mother & 56 &  31 &  25 &  1 &  0 &  55  \\
\cline{2-8}
& Father & 14 &  11 &  3 & 1 &  1 & 12  \\
\cline{2-8}
& Investigator & 15 &  10 &  5 & $\uparrow$ 2 &  1 &  $\downarrow$ 12  \\
\hline
\hline
\multirow{4}{*}{Verbs}
& Children & 56 & $\uparrow$ 35 & $\downarrow$ 21 &  $\uparrow$ 6 &  1 & $\downarrow$ 49  \\
\cline{2-8}
& Mother & 56 &  31 &  25 &  1 & $\uparrow$ 5 & $\downarrow$ 50  \\
\cline{2-8}
& Father & 14 &  7 &  7 &  0 &  0 &  14  \\
\cline{2-8}
& Investigator & 15 &  8 &  7 & $\uparrow$ 2 &  0 &  13  \\
\hline
\hline
\multirow{4}{*}{Adjectives}
& Children & 54 & $\uparrow$ 35 & $\downarrow$ 19 &  $\uparrow$ 7 &  2 &  $\downarrow$ 45  \\
\cline{2-8}
& Mother & 55 &  25 &  30 &  0 & 3 &  52  \\
\cline{2-8}
& Father & 13 &  5 &  8 &  0 &  1 &  12  \\
\cline{2-8}
& Investigator & 15 &  7 &  8 &  0 & $ \uparrow$ 2 & 13  \\
\hline
\hline
\multirow{4}{*}{Adverbs}
& Children & 56 &  28 &  28 &  $\uparrow$ 5 &  $\uparrow$ 5 & $\downarrow$ 46  \\
\cline{2-8}
& Mother & 56 & $\uparrow$ 37 & $\downarrow$ 19 & $\uparrow$ 4 &  1 & 51 \\
\cline{2-8}
& Father & 14 &  8 &  6 &  0 &  0 &  14  \\
\cline{2-8}
& Investigator & 15 &  5 &  10 &  1 &  0 &  14  \\
\hline
\end{tabular}
\caption{Raw data on the evolution of \textbf{SemCor polysemy} over
	time taking into account the syntactic category and role.
	The format and sample length are the same as
	in Table \ref{tab:eng_edat_sinsets_binom_50}.
}
\label{tab:eng_edat_sinsets_semcor_binom_50}
\end{center}
\end{table}
}

\def\taulaDOS{
\begin{table}[hp!]
	\begin{center}
		\begin{tabular}{|l|c|r|r|c||r|r|}
			\hline
			\textbf{Role} & \textbf{POS} & \textbf{Mean} & \textbf{Dev} & \textbf{Max} & \textbf{Tokens} & \textbf{\%} \\
			\hline
			\hline
			\multirow{4}{*}{Children}
			& Adjective & 6.01 & 5.87 & 28 & 90,512 & 6.66 \\
			\cline{2-7}
			& Adverb    & 2.71 & 2.77 & 16 & 211,953 & 15.61 \\
			\cline{2-7}
			& Noun      & 3.65 & 3.71 & 33 & 588,582 & 43.33 \\
			\cline{2-7}
			& Verb      & 15.29 & 11.45 & 59 & 467,172 & 34.40 \\
			\hline
			\hline
			\multirow{4}{*}{Mother}
			& Adjective & 6.30 & 6.11 & 28 & 172,685 & 7.61 \\
			\cline{2-7}
			& Adverb    & 3.09 & 3.39 & 16 & 354,798 & 15.63 \\
			\cline{2-7}
			& Noun      & 3.84 & 3.89 & 33 & 742,693 & 32.72 \\
			\cline{2-7}
			& Verb      & 14.48 & 10.15 & 59 & 999,625 & 44.04 \\
			\hline
			\hline
			\multirow{4}{*}{Father}
			& Adjective & 6.84 & 6.62 & 28 & 28,299 & 9.02 \\
			\cline{2-7}
			& Adverb    & 3.33 & 3.47 & 16 & 49,187 & 15.68 \\
			\cline{2-7}
			& Noun      & 4.01 & 4.05 & 33 & 101,111 & 32.24 \\
			\cline{2-7}
			& Verb      & 14.15 & 10.08 & 59 & 134,996 & 43.05 \\
			\hline
			\hline
			\multirow{4}{*}{Investigator}
			& Adjective & 7.02 & 6.62 & 28 & 13,509 & 7.41 \\
			\cline{2-7}
			& Adverb    & 3.18 & 3.35 & 16 & 29,409 & 16.12 \\
			\cline{2-7}
			& Noun      & 4.15 & 3.90 & 33 & 56,534 & 30.99 \\
			\cline{2-7}
			& Verb      & 13.94 & 9.91 & 59 & 82,950 & 45.48 \\
			\hline
		\end{tabular}
		\caption{Statistics about the number of synsets of tokens by role
		(children, mother, father and investigator) and part-of-speech in CHILDES: mean ($Mean$),  standard deviation ($Dev$) and maximum number of synsets ($Max$). $Tokens$ is the total number of tokens.
			$\%$ is the percentage of each part-of-speech over the total number of tokens.}

		\label{tab:syn-childes}
	\end{center}
\end{table}
}


\newcommand{\taulaTestAnovaWordnet}{
\begin{center}
\begin{table}[H]
\resizebox{\columnwidth}{!}{
\begin{tabular}{|c||c|c|c|c|c|c|c|c|c|c|}
\hline
\diaghead{Scoreexp}{from}{to} &
\diaghead{Scoreexp}{ 11.9 }{ 17.4 } &
\diaghead{Scoreexp}{ 17.4 }{ 22.9 } &
\diaghead{Scoreexp}{ 22.9 }{ 28.4 } &
\diaghead{Scoreexp}{ 28.4 }{ 33.9 } &
\diaghead{Scoreexp}{ 33.9 }{ 39.4 } &
\diaghead{Scoreexp}{ 39.4 }{ 44.9 } &
\diaghead{Scoreexp}{ 44.9 }{ 50.3 } &
\diaghead{Scoreexp}{ 50.3 }{ 55.8 } &
\diaghead{Scoreexp}{ 55.8 }{ 61.3 } &
\diaghead{Scoreexp}{ 61.3 }{ 66.8 } \\
\hline \hline

N & 14 & 99 & 139 & 139 & 127 & 96 & 25 & 18 & 48 & 9 \\
\hline

F & \textbf{13.331} & \textbf{158.401} & \textbf{90.728} & \textbf{35.397} & \textbf{11.723} & \textbf{7.659} & \textbf{3.827} & 1.751 & 0.859 & 0.005 \\
\hline

p-val & 0.002 & $ < 10^{-21}$ & $ < 10^{-16}$ & $ < 10^{-7}$ & $ < 10^{-3}$ & 0.003 & 0.031 & 0.102 & 0.179 & 0.473 \\
\hline

\end{tabular}
}
\caption{A comparison of the mean WordNet polysemy of children versus that of adults.
The analysis is based on a one-way independent samples ANOVA with
	\textit{role} (two levels: children and adults) as fixed factor
	and the \textit{identifier} of the individual as random factor.
	The test is performed in 10 segments of the total timeline.
	 \textbf{from-to} indicates the first and last months of the
	$i$-th segment of time. \textbf{N} is the size of the sample.
	\textbf{F} is the F-value of the Anova test.
	\textbf{p-val} is the p-value of the test.
	The F-values in \textbf{bold} are significant.}
\label{testAOVwordnet}
\end{table}
\end{center}
}


\newcommand{\taulaTestAnovaSemcor}{
\begin{center}
\begin{table}[H]
\resizebox{\columnwidth}{!}{
\begin{tabular}{|c||c|c|c|c|c|c|c|c|c|c|}
\hline
\diaghead{Scoreexp}{from}{to} &
\diaghead{Scoreexp}{ 11.9 }{ 17.4 } &
\diaghead{Scoreexp}{ 17.4 }{ 22.9 } &
\diaghead{Scoreexp}{ 22.9 }{ 28.4 } &
\diaghead{Scoreexp}{ 28.4 }{ 33.9 } &
\diaghead{Scoreexp}{ 33.9 }{ 39.4 } &
\diaghead{Scoreexp}{ 39.4 }{ 44.9 } &
\diaghead{Scoreexp}{ 44.9 }{ 50.3 } &
\diaghead{Scoreexp}{ 50.3 }{ 55.8 } &
\diaghead{Scoreexp}{ 55.8 }{ 61.3 } &
\diaghead{Scoreexp}{ 61.3 }{ 66.8 } \\
\hline \hline

N & 14 & 99 & 139 & 139 & 127 & 96 & 25 & 18 & 48 & 9 \\
\hline

F & \textbf{18.141} & \textbf{241.664} & \textbf{173.963} & \textbf{82.031} & \textbf{31.37} & \textbf{15.817} & \textbf{12.408} & \textbf{8.731} & \textbf{3.313} & 0.602 \\
\hline

p-val & $ < 10^{-3}$ & $ < 10^{-27}$ & $ < 10^{-25}$ & $ < 10^{-15}$ & $ < 10^{-7}$ & $ < 10^{-4}$ & $ < 10^{-3}$ & 0.005 & 0.038 & 0.232 \\
\hline

\end{tabular}
}
\caption{The same as in Table \ref{testAOVwordnet}, here for SemCor polysemy. }
\label{testAOVsemcor}
\end{table}
\end{center}
}


\newcommand{\taulaTestAW}{
\begin{tabular}{|c|c||c|c||c|c|}
\multicolumn{2}{c}{} & \multicolumn{2}{c}{WordNet} & \multicolumn{2}{c}{SemCor} \\
\hline 
\diaghead{Scoreexp}{from}{to} & N & F & p-val & F & p-val \\
\hline \hline
\diaghead{Scoreexp}{ 11.9 }{ 17.4 } & 14 & \textbf{13.331} & 0.002 & \textbf{18.141} & $ < 10^{-3}$ \\
\hline
\diaghead{Scoreexp}{ 17.4 }{ 22.9 } & 99 & \textbf{158.401} & $ < 10^{-21}$ & \textbf{241.664} & $ < 10^{-27}$ \\
\hline
\diaghead{Scoreexp}{ 22.9 }{ 28.4 } & 139 & \textbf{90.728} & $ < 10^{-16}$ & \textbf{173.963} & $ < 10^{-25}$ \\
\hline
\diaghead{Scoreexp}{ 28.4 }{ 33.9 } & 139 & \textbf{35.397} & $ < 10^{-7}$ & \textbf{82.031} & $ < 10^{-15}$ \\
\hline
\diaghead{Scoreexp}{ 33.9 }{ 39.4 } & 127 & \textbf{11.723} & $ < 10^{-3}$ & \textbf{31.37} & $ < 10^{-7}$ \\
\hline
\diaghead{Scoreexp}{ 39.4 }{ 44.9 } & 96 & \textbf{7.659} & 0.003 & \textbf{15.817} & $ < 10^{-4}$ \\
\hline
\diaghead{Scoreexp}{ 44.9 }{ 50.3 } & 25 & \textbf{3.827} & 0.031 & \textbf{12.408} & $ < 10^{-3}$ \\
\hline
\diaghead{Scoreexp}{ 50.3 }{ 55.8 } & 18 & 1.751 & 0.102 & \textbf{8.731} & 0.005 \\
\hline
\diaghead{Scoreexp}{ 55.8 }{ 61.3 } & 48 & 0.859 & 0.179 & \textbf{3.313} & 0.038 \\
\hline
\diaghead{Scoreexp}{ 61.3 }{ 66.8 } & 9 & 0.005 & 0.473 & 0.602 & 0.232 \\
\hline
\end{tabular}
}


\newcommand{\taulaTestAnovaWordnetX}{
\begin{center}
\begin{table}
\centering
\taulaTestAW 
\caption{A comparison of the mean WordNet and SemCor polysemies 
of children versus that of adults.
The analysis is based on a one-way independent samples ANOVA with
\textit{role} (two levels: children and adults) as fixed factor
and the \textit{identifier} of the individual as random factor.
The test is performed in 10 segments of the total timeline.
 \textbf{from-to} indicates the first and last months of the
$i$-th segment of time. \textbf{N} is the size of the sample.
\textbf{F} is the F-value of the Anova test.
\textbf{p-val} is the p-value of the test.
The F-values in \textbf{bold} are significant.}
\label{testAOVwordnetX}
\end{table}
\end{center}
}

\thispagestyle{headings}
\pagenumbering{arabic}

\section*{Abstract}

\begin{quotation}

Here we study polysemy as a potential learning bias
in vocabulary learning in children. Words of low polysemy 
could be preferred as they reduce the disambiguation effort for the listener.
However, such preference could be a side-effect of another bias:
the preference of children for nouns in combination with the lower polysemy of nouns
with respect to other part-of-speech categories.

Our results show that mean polysemy in children increases
over time in two phases, i.e. a fast growth till the
31st month followed by a slower tendency towards adult speech.
In contrast, this evolution is not found in adults interacting with children.
This suggests that children have a preference
for non-polysemous words in their early stages of vocabulary acquisition.
Interestingly, the evolutionary pattern described above weakens 
when controlling for syntactic category
(noun, verb, adjective or adverb) but it does not disappear completely, 
suggesting that it could result from a
combination of a standalone bias for low polysemy and a preference for nouns.

\bigskip

\emph{Keywords:} {
child language evolution,
vocabulary learning,
learning biases,
polysemy,
quantitative linguistics
}
\end{quotation}

\section{Introduction}
\label{sec:introduction}

Children are exposed to millions of word tokens through an accumulation of small
interactions grounded in context, immersed in a \textit{sea of words} that
they learn early after their birth \parencite{Royetal2015}.
Infants begin learning their native language by discovering
aspects of its sound structure, eventually learning words
\parencite{Swingley2009} after exposure to their mother tongue.
Initially, ``knowing a word'' requires knowing the sound form
of the word and its speech segmentation (i.e. by infants'
use of subtle prosodic boundary markers [\cite{Gout2004}])
and denotation, including its semantic reference and other
syntactic properties, so infants younger than 12 months
may know very few words but in general know perfectly
the sounds and general mechanisms of segmentation of
their mother tongue ([\cite{Swingley2009}] for a review).

In fact, human language development follows a predictable sequence that
traditionally has led linguists and psychologists
to establish some phases, or stages, in the process of language acquisition
\parencite{Brown1973}, with well-known critical periods \parencite{Kuhl2010}.
At present we cannot determine exactly the timing of these critical periods
for each individual, but many studies
suggest the existence of a critical period for syntactic learning approximately
between 18 and 36 months of age preceded
by a critical period for the phonemic level prior to 12 months \parencite{Kuhl2010, Yule2006}.

During this process, some words are learned first instead of others and many biases
have been hypothesized 
in order to explain why some words are learned earlier
\parencite{Saxton10childlanguage}.
For instance, a preference for
\begin{seriate}
\item basic taxonomic level, i.e. less generic words \parencite{Mervis1987a, Tomasello2000b},

\item nouns, e.g., children learn first nouns and later verbs
	  \parencite{Gentner1982, Gentner2006, McDonough2011},

\item contextual distinctiveness
	  in space, time or linguistic environment \parencite{Royetal2015},

\item frequent words \parencite{JCL:1917448},

\item neighborhood density, i.e. number of words that sound similar
	to a given word \parencite{Storkel2004a,Storkel2006a},

\item in-degree of the word in free association norms \parencite{Hills2009a, Hills2010a} or

\item mutual information when associating a new word to a meaning \parencite{Ferrer2013g}.
\end{seriate}

Actually, the interest in vocabulary learning biases goes 
beyond child language research:
it has been hypothesized that early stages of language have left traces
of simple forms of language \parencite{Bickerton1990,Jackendoff1999a}, 
child language being one example.
On the one hand, word learning biases in children suggest 
constraints that human language
faced at its very origin \parencite{Saxton10childlanguage}. 
On the other hand,
the factors by which these learning biases are overridden suggest 
processes by which communication systems
achieve higher linguistic complexity \parencite{Saxton10childlanguage}.

In addition to the biases listed above, here we explore the existence
of a key and polyhedral bias in language acquisition:
the preference for less polysemous words.
In a broad sense, we may define polysemy as the capacity
of a word form to have more than one meaning since the pioneering work of Br\'eal,
who considered polysemy as a diachronic phenomenon with synchronic effects:
words acquire new meanings through use, 
which coexist with the old ones \parencite{Breal1897}. 
In fact, he observed that the communication context
determines the sense of a polysemous word and
eliminates the ambiguity for listeners. 
However, despite being polysemy an object of
study for more than a hundred years, there are terminological differences
in its definition (and other related terms) depending on the approach
\parencite{Kovacs2011}.

Language researchers usually use the concept of \textit{ambiguity},
which is split into 
\textit{polysemy} (in a narrow sense), 
in case of word forms with multiple related meanings
and 
\textit{homonymy}, in case of words forms with multiple unrelated meanings,
\parencite{Rodd2002a}.
This definition of polysemy narrows our definition above because
it requires that the multiple meanings of a word form must be related.
An example of polysemy
is the word \emph{walk}, that has related meanings such as \texttt{go for a walk}
or \texttt{walk the dog}, among others \parencite{Preda2013}.
A classical example of homonymy is the word \emph{bank}, that has apparently
unrelated meanings in English such as \texttt{a financial institution for storing money}
or \texttt{the land alongside or sloping down to a river} 
\parencite{Lyons1982a,Allan1986a,Langone2004a}.
However, the view of these meanings as unrelated is not conclusive.
In fact, a \texttt{bank} (financial institution) was originally
the table that medieval money dealers used,
and it still has that meaning in some Romance languages (e.g. Catalan and Spanish),
although many speakers of these languages are unaware of this fact.
This raises the question of whether one should 
consider the definitions of polysemy (in a narrow sense) or homonymy,
depending on the language, the speaker's knowledge,
or even the associated neuronal activity \parencite{Klepous_etal_2012}.
This is a clear example of the complexity of the distinction between polysemy
and homonymy that is beyond the scope of the present article, but yet, 
a hot topic of research in Linguistics and Neuroscience.

Language researchers also distinguish between homophony and homography.
Two words are homophones if they match phonetically but differ in meaning,
like in the pair \emph{be/bee}, whereas two words are homographs
if they have the same orthographical representation but differ in meaning \parencite{trask1996}.
Due to the nature of our data, our definition of polysemy 
collapses the definition of polysemy in a narrow sense and homography (not homophony).
Put differently, here we will investigate a preference for less polysemous
word types as defined according to their written form, i.e.
independently from their pronunciation, and understanding polysemy in a broad sense.

The potential preference for non-polysemous words
suggested above may be:

\begin{APAitemize}
	\item \textbf{a standalone bias}.
	This hypothesis would be consistent with the lower uncertainty
	for those words with respect to their meaning,
	a factor that may reduce the cognitive cost of learning them
	as the cost for the listener would be smaller \parencite{Zipf1949a}.

	\item \textbf{a side-effect} of another bias,
	for instance an initial preference for nouns,
	the so-called \textsl{noun-bias}
	\parencite{Gentner1982, Gentner2006, McDonough2011},
	combined with the fact that nouns have lower polysemy \parencite{Fausey_etal_2006}.

On the one hand, a standalone preference for low polysemy
goes back to Zipf's principle of Least Effort on the side of listener \parencite{Zipf1949a}:
the cost of determining the actual meaning is expected to be smaller when there
are less candidate meanings. However, \textcite{Ullmann1959} considered that
\textit{polysemy is an indispensable resource of language economy} \parencite{Kovacs2011},
suggesting a pressure in the opposite direction. It would be altogether impracticable
to have separate terms for every referent \parencite{Ferrer2003a,Kovacs2011},
in the spirit of Zipf's effort for the speaker \parencite{Zipf1949a}.
On the other hand, the acquisition of words according to their syntactic category
has been studied deeply by researchers. The acquisition of verb polysemy
has become an important and productive target of study and modeling in linguistics
and psychology \parencite{ParisienStevenson2009}. Although some of the most frequent
and earliest learned English verbs are also among those with the largest number of
senses \parencite{Clark1996}, in general, nouns tend to appear earlier than verbs
and a possible interpretation is that nouns are generally easier to learn than verbs
in all languages \parencite{Gentner1982}. In the first year of life infants are able
to form mental representations of both objects and events, they notice
the connection between objects and actions in which they were engaged and,
consequently, by 24 months, infants tend to map novel nouns to objects and novel
verbs to actions, certainly under a social and cultural influence \parencite{Waxman_etal2016}.
However, neuroscience is just at the beginning of revealing the brain systems
that underlie the human language faculty \parencite{Kuhl2010} and
this traditional view is still under discussion \parencite{Gogate_Hollich_2017}.

When exploring biases in child speech, it is crucial to distinguish
between biases which are genuine and biases which may just be the result
of mimicking some form of adult speech.
This bring us to the question
of the so-called \emph{motherese} or \emph{infant-directed speech} (IDS),
or how adults talk to kids, that may make word learning and comprehension easier
for infants, especially in the first months \parencite{Foursha-Stevenson2017},
or at least some tasks \parencite{Thiessen2005}. This has important epistemological
implications in studies of language acquisition, because if we take as usual adult
speech as control, considering that adult speech is the target stage 
in the evolution of the verbal production of a child, then it is crucial, especially
for quantitative studies, to know what kind of input is receiving the child and
if the adult is performing adaptive efforts (aimed at facilitating
the learning of their offspring) or not. Also, imitation is a powerful form of
learning commonly used by children not only in language \parencite{Meltzoff2003},
because infants are born to learn, and they learn at first by imitating us \parencite{Meltzoff_1999}.
Therefore, in short, we also have to involve the linguistic input that
children receive over time in our analyses.

The issue of infant-directed speech requires a statistical framing.
The null hypothesis is that children speak by sampling uniformly at random over
the word tokens in the form of adult speech they receive.
Intuitively, this null hypothesis means that children are merely mirroring child-directed speech.
If the null hypothesis was true, children should be showing exactly the same statistical
characteristics of adult speech. The alternative hypothesis is that children are
producing words with some bias with respect to adult speech. If the null hypothesis did not hold,
the challenge would be to shed light on the nature of that bias.

We analyze here the variation in mean polysemy in children over time with the help of a
massive database that contains the transcriptions of conversations between children and adults
(CHILDES Database [\cite{MacWhinney2000a}]) and using words as our unit of study 
(understood as written words,
separated by blanks, as usual in corpus-based computational linguistics).
Consequently we do not consider here other theoretical units (i.e. constructions)
common in some approaches to study language acquisition \parencite{Diessel2013}.
The period of time analyzed (child age) runs from 10 to 60 months. 
We use two measures of polysemy: the total number of meanings of a 
word according to the WordNet lexical database, and the number of
WordNet meanings of a word that have appeared in an annotated corpus (the SemCor corpus).
For further details, see Section \nameref{sec:materialsandmethods}.
These two corpora allow us to take into account the difference 
between productive and potential vocabulary knowledge.
One general limitation in Computational Linguistics 
(that we obviously assume here) is that we study what children can produce, 
i.e. their productive vocabulary,
but not what they can understand, i.e. their receptive vocabulary \parencite{Schmitt2008}.

We will show that mean polysemy in children presents a pattern
which is significantly different from that of adults:
the mean polysemy in adults remains practically stable during all
the analyzed timespan, whereas that of children
starts with a low value, and increases rapidly
to almost converge with the mean polysemy of adults.
Finally we consider two plausible scenarios to explain our results:
a standalone bias or the preference for a specific syntactic category.

\end{APAitemize}

\section{Materials and Methods}
\label{sec:materialsandmethods}

\subsection{CHILDES database}
\label{sub:childes}

The longitudinal studies of child language development were taken from
the CHILDES database \parencite{MacWhinney2000a}.
The majority of corpora within this database are transcripts of
conversational interactions among children and adults
which occur at a given point in time, which are referred to as
\textbf{recording sessions} throughout this paper.
\textcolor{black}{
Besides children, adults are included to control for the possible effect of child-directed speech,
the so-called \emph{motherese} (see Section \nameref{sec:introduction}).
The period of time analyzed (child age) runs from 10 to 60 months.
The distribution of time points and the production
per time point is not homogeneous.}

The longitudinal studies of child language development from
the CHILDES database
that were analyzed are the same as in \parencite{Baixeries2012c} for English language: 60 target children.
In this article we only analyze content words (nouns, verbs, adjectives and adverbs).
All corpora of the CHILDES database are freely available at
\url{https://childes.talkbank.org/}
(accessed 17 December 2012).

\subsection{WordNet lexical database}
\label{lexical_databases_section}

The polysemy of a word in English was obtained by querying the lexical
database WordNet,
that can be seen as a set of synsets and relationships among them \parencite{Miller1995a, FellBaum1998a}.
A synset is the representation of an abstract meaning and is defined as a
set of words having (at least) the meaning that the synset stands for.
As an example in WordNet, the word \emph{book} is related (among others) with the synset
that represents \textit{a written work or composition that has been published},
with the synset that represents \textit{a written version of a play or other
	dramatic composition; used in preparing for a performance} or the synset
that represents \textit{to arrange for and reserve (something for someone else) in advance}.
Each pair word-synset contains, as well, the information that corresponds
to a syntactic category related to those two elements.
For instance, the pair \emph{book} and the synset
\textit{a written work or composition that has been published}
are related to the syntactic category \emph{noun}, whereas the
pair \emph{book} and synset
\textit{to arrange for and reserve (something for someone else) in advance}
are related to the syntactic category \textit{verb}.

WordNet has 155,287 lemmas and 117,659 synsets (see \url{http://wordnet.princeton.edu/wordnet/man/wnstats.7WN.html}), and contains only
four main syntactic categories: nouns, verbs,
adjectives and adverbs. Words with other syntactic categories are not present in this
database (for instance, the article \emph{the} or the preposition \emph{for}).

WordNet is freely available for download at
\url{http://wordnet.princeton.edu/wordnet/download/}.
We use WordNet version 3.0, that it is included in the NLTK platform
(Natural Language Toolkit) version 2.0 freely available at \url{http://www.nltk.org}.

\subsection{SemCor corpus}

The Semantic Concordance Package (SemCor) is a corpus 
composed of 352 texts which are a subset of the English Brown Corpus.
All words in the corpus were syntactically tagged using
Brill's part-of-speech tagger.
The semantic tagging was done manually,
mapping all content words
to their corresponding synsets in WordNet.

SemCor contains $676,546$ tokens, $234,136$ of which are sense-tagged. 
This yields $23,341$ different tagged lemmas that represent only content words.

We processed SemCor to count the number of different WordNet meanings
that every pair <lemma, syntactic category> has in this corpus.
All this information was saved to be used in the processing of
the recording sessions.

Table \ref{fig:stats-semcor} shows the percentage of CHILDES 
lemmas that are present in SemCor corpus.
The coverage of SemCor (proportion of SemCor lemmas that appear in WordNet) for verbs,
adjectives and adverbs is above 75\%, for nouns it is lower, between 53,61\% and 59,32\%.

\begin{table}[htp!]
\begin{center}
\begin{tabular}{|l|c|c|c|c||c|c|c|}
\hline
\textbf{Role} & \textbf{POS} & \textbf{\# typ WN} & \textbf{\# typ SC} & \textbf{\% typ} & \textbf{\# tok WN} & \textbf{\# tok SC} & \textbf{\% tok}\\
\hline
\hline
\multirow{4}{1.5cm}{Children} & n & 5,973 & 2,616 & 43.80 & 588,582 & 315,518 & 53.61\\
& v & 1,584 & 1,109 & 70.01 & 467,172 & 443,555 & 94.94\\
& a & 1,261 & 731 & 58.05 & 90,512 & 76,102 & 84.08 \\
& r & 369 & 246 & 66.66 & 211,953 & 159,094 & 75.06 \\
\hline
\hline
\multirow{4}{1.5cm}{Mother} & n & 8,023 & 3,571 & 44.51 & 742,693 & 439,839 & 59.22\\
& v & 2,438 & 1,738 & 71.29 & 999,625 & 965,200 & 96.56\\
& a & 2,131 & 1,251 & 58.70 & 172,685 & 155,212 & 89.88\\
& r & 633 & 436 & 68.88 & 354,798 & 311,011 & 87.66\\
\hline
\hline
\multirow{4}{1.5cm}{Father} & n & 4,172 & 2,279 & 54.63 & 101,111 & 58,040 & 57.40 \\
& v & 1,247 & 992 & 79.55 & 134,996 & 130,698 & 96.82\\
& a & 1,013 & 671 & 66.24 & 28,299 & 26,202 & 92.59\\
& r & 376 & 279	& 74.20 & 49,187 & 42,039 & 85.47\\
\hline
\hline
\multirow{4}{1.5cm}{Investig.} & n & 2,361 & 1,490 & 63.11 & 56,534 & 33,538 & 59.32\\
& v & 785 & 658	& 83.82 & 82,950 & 79,974 & 96.41\\
& a & 577 & 439 & 76.08 & 13,509 & 12,753 & 94.40\\
& r & 254 & 206 & 81.10 & 29,409 & 24,868 & 84.56\\
\hline
\end{tabular}
\end{center}
\caption{Statistics of processed lemmas of CHILDES that appear in SemCor ($SC$) by syntactic category according to role
(Children, Mother, Father and Investigator).
\emph{POS}: Part-of-speech (n: nouns, v: verbs, a: adjectives, r: adverbs).
\emph{\# typ\ WN}: number of types that appear in WordNet. 
\emph{\# typ\ SC}: number of types that appear in SemCor.
\emph{\% typ}: percentage of types in WordNet that appear in SemCor. 
\emph{\# tok\ WN}: number of tokens in WordNet.
\emph{\# tok\ SC}: number of tokens that appear in SemCor. 
\emph{\% tok}: percentage of tokens in WordNet that appear in SemCor.}
\label{fig:stats-semcor}
\end{table}

We use SemCor version 3.0, freely available for download at
\url{http://web.eecs.umich.edu/~mihalcea/downloads.html#semcor}.

\subsection{TreeTagger}

TreeTagger \parencite{treetagger1994} was used to determine 
the syntactic category and the lemma of word
tokens from transcripts of speech.
Essentially, this tool annotates text with part-of-speech (POS) 
and lemma information (canonical form).
That tool was chosen because it supports many languages,
English among others, which could facilitate the extension 
of our studies to other languages as well in the future.

\DefineShortVerb{\_}

In English, TreeTagger POS tags that refer to content words are:
\begin{itemize}
	\item Adjectives: _JJ_, _JJR_ and _JJS_.
	\item Adverbs: _RB_, _RBR_ and _RBS_.
	\item Nouns: _NN_ and _NNS_.
	\item Verbs: _MD_, _VBD_, _VBG_, _VBN_, _VBP_, _VBZ_, _VD_, _VDD_, _VDG_, _VDN_, _VDP_, _VDZ_, _VH_, _VHD_, _VHG_, _VHN_, _VHP_, _VHZ_, _VV_, _VVD_, _VVG_, _VVN_, _VVP_ and _VVZ_ .
\end{itemize}

\UndefineShortVerb{\_}

\subsection{Processing data}

For each recording session, we processed the list with all tokens of the
transcript in the same order that have been produced by a certain speaker.
For each of these tokens, we proceeded as follows:
\begin{APAenumerate}
	
	\item \emph{Tagging}. Each token in this list was assigned a morpho-syntactic category tag
	and a lemma by TreeTagger. Therefore, this list contained now triples
	<token, syntactic category, lemma>.

	\item \emph{Discard unprocessable tokens}. We discarded from the previous list those
	triples whose tokens
	have non-ascii chars 
	or are CHILDES special tags ("@", "xxx", "xx", "yyy", "yy", "www").
	
	\item \emph{Proper noun recognition}. We recognized the proper nouns
	from a list of proper nouns. Any triple whose token
	appeared in this list was tagged as proper noun.
	This list was compiled with information extracted from different sources and
	can be downloaded at \url{http://tinyurl.com/polysemy-ne-txt}.

	\item \emph{Category filtering}.
	From the resulting list in the previous step, we only selected those triples whose
	morpho-syntactic category corresponds to a content word.

	\item \emph{Polysemy calculation}. We used two methods to calculate the polysemy
	for every <token, syntactic category, lemma> of the list computed in the previous step:
	\begin{quotation}
	\begin{APAitemize}
	
	\item \emph{WordNet query}.
		WordNet was queried to obtain the number of synsets related to each triple
		<token, syntactic category, lemma> from the previous list.
		We first queried WordNet about the pair <token, syntactic category>.
		If this pair was not found, we queried again WordNet about the pair
		<lemma, syntactic category>.
		If this pair was not found, this token was discarded
		and not considered for calculations of WordNet polysemy.

	\item \emph{SemCor query}.
		The file with the processed information of SemCor was queried
		to obtain the number of synsets related to each triple
		<token, syntactic category, lemma> from the previous list.
		We queried WordNet about the pair <lemma, syntactic category>.
		If this pair was not found, this token was discarded
		and not considered for calculations of the SemCor polysemy.

	\end{APAitemize}
	\end{quotation}
\end{APAenumerate}

\textcolor{black}{The percentage of discarded tokens after applying all filters is:
	children 50.70\%, mother 51.56\%, father 50.73\% and investigator 51.85\%. }

For instance, in the example above where the token \emph{book}
was tagged to be a noun, we would only select the 11 synsets
yielded by WordNet for the \emph{noun} category.
If there were no synsets related to <\emph{book}, noun>
this token would be discarded.

We also performed some extra controls:

\begin{APAenumerate}

\item A \textbf{control for sample length}:
from the remaining list that have passed the previous
filters, we selected the first $n$ tokens
(where $n = 50, 75, 100, 125, 150, 175, 250$) for each recording session.
If the sample did not have at least $n$ tokens then the entire
transcript was discarded.

\item A \textbf{control for morpho-syntactic category}:
from the remaining list, we only
considered those tokens that belong to a specific morpho-syntactic category:
noun, adjective, verb or adverb.
In the study of the specific morpho-syntactic category
we also controlled for sample length:
we took the first $n$ tokens
after applying the morpho-syntactic category filter
(tokens that do not belong to the target syntactic category are discarded).
As before, if the sample did not have, at least, $n$ tokens, then, the entire
transcript was discarded.

\end{APAenumerate}

Finally, notice that all the processing steps above imply that the analysis
that is not restricted to a specific morpho-syntactic category
defines the polysemy of a token according to the morpho-syntactic category of the triple.
It does not define the polysemy aggregating the polysemy
for all syntactic categories somehow (e.g., by summing the polysemy
of the token for all syntactic categories). The reason for our choice is
to run the raw analysis in as similar a way as possible to that carried out
when controlling for morpho-syntactic category.

\subsection{Mathematical computations}
\label{subsec:mathematical}

The association between mean polysemy and age was measured 
with a Spearman rank correlation,
$\rho_S(X,Y)$, which is a measure of monotonic dependency between a pair of variables
\parencite{Conover1999a}.
Although the traditional Pearson correlation has been used to investigate
the relationship between polysemy and time in L2 learners \parencite{Crossley2010a},
Spearman rank correlation has the advantage of being able 
to capture non-linear dependencies;
the traditional Pearson correlation is simply a measure of 
linear dependency \parencite{Gibbons2010a,Embrecths2002a}.

To determine whether a correlation was significant or not, a two-sided
correlation test with a significance level of $\alpha = 0.05$ was used (a
correlation is significant if the p-value does not exceed $\alpha$).

The smoothing of the plots in Figures \ref{fig:eng_edat_sinsets_all_50},
\ref{fig:eng_edat_sinsets_other_50} and \ref{fig:eng_edat_category}
was performed with the non parametric smoother method \verb!lowess!
implementated in Python in the library \emph{statsmodels}
(see \url{http://statsmodels.sourceforge.net/devel/generated/statsmodels.nonparametric.smoothers_lowess.lowess.html})
with parameters $frac= 1./3, it=0$.

All participants with less than $m^*$ time points were excluded from the analyses.
$m^*$ is the minimum number of points that are needed for significance by a two-sided correlation
test between two vectors $X$ and $Y$. $m^*$ is the smallest value of $m$ satisfying the condition
\parencite{Ferrer2012a}
\begin{equation}
2/(m!) \leq \alpha,
\end{equation} 
where $\alpha$ is the significance level.
With $\alpha = 0.05$ then $m^* = 5$.

\subsection{Fisher method of randomization}
\label{subsec:fisher}

To assess whether verbs have a higher mean polysemy than
other syntactic categories for a given role, we performed
a Fisher randomization test \parencite{Conover1999a}.
The statistic used is the absolute value of the difference between
the mean polysemy of the verb tokens and the mean polysemy of the tokens of the other category.
The test checks whether the absolute difference is significantly high.
The p-value of the test was determined by means of a Monte Carlo procedure over $10^5$ randomizations.

\subsection{Binomial tests}
\label{subsec:binomialtest}

The $\uparrow$ and $\downarrow$ arrows in Tables \ref{tab:eng_edat_sinsets_binom_50}
and \ref{tab:eng_edat_sinsets_semcor_binom_50} indicate
if a certain number is significantly high or significantly low.
These arrows were determined with the help of binomial tests \parencite{Cross1982a,Baixeries2012c}.
Suppose that $N$ is the number of individuals with at least $m^*$ points of time
and $\alpha$ is the significance level of that test. Under the null hypothesis,

\begin{itemize}
	\item $C_+$ and $C_-$ follow a binomial distribution with parameters $N$ and $1/2$.

	\item $S_-$ and $S_+$ follow approximately a binomial distribution with parameters $N$ and $\alpha/2$.

	\item $S_?$ follows approximately a binomial distribution with parameters $N$ and $1-a$. 
\end{itemize}

\subsection{Breakpoint calculation}
\label{subsec:calculationbreakpoint}

A breakpoint in the relationship between time and another variable
(e.g., mean polysemy)
was computed using the R package \emph{strucchange}
\parencite{strucchange, Zeileis2003109} assuming that the curves 
contain a single breakpoint (setting the parameter {\em breaks} to 1).
This R package implements the algorithm described
by \textcite{BaiPerron2003}
for simultaneous estimation of multiple breakpoints,
but in our case we are just looking for one breakpoint.
The  breakpoint is the value of the predictor (time or children's age in our case)
where the error is minimized in a double linear regression fit.
In all cases, we get breakpoints with an error of $\pm 0.1$ months 
according to a confidence level of $95\%$.

\subsection{Anova test computation}
\label{subsec:anovatestcomputation}

\textcolor{black}{We have applied an Anova test to ten different segments
of equal size of the timeline.
In each segment, we have averaged the mean polysemy for each individual,
which is identified with a distinct identifier.
We have also grouped all adults (mothers, fathers and investigators)
into one single role.
We perform a one-way independent samples Anova test
with a fixed factor \textit{role} (with two different levels:
target children and adults), and the random factor of this test was the
identifier for each individual.
Our alternative hypothesis is that the average
(both WordNet and SemCor) polysemy
for children is significantly lower than that of adults. 
Thus we have performed a one-tailed test with the 
help of the two-tailed test provided by the function
{\tt aov} of the standard R library {\tt stats}. The p-value returned by {\tt aov}
is halved to obtain the p-value of the one-tailed test. }

\clearpage

\section{Results}
\label{sec:results}

\subsection{Evolution of the polysemy over time}

We estimate the polysemy of a speaker e.g., a child,
from a continuous speech sample,
which can be seen as a sequence of $N$
word tokens, $t_1,...,t_i,...t_N$.
The \textbf{mean polysemy} of a sample is defined as the mean number of meanings of the $i$-th token
according to its part-of-speech
i.e. noun, adjective, adverb or verb (in these samples, only content words are considered).
Samples were obtained from transcripts of recording sessions
from the CHILDES database \parencite{MacWhinney2000a}
of speech between children and adults 
(see Section \nameref{sec:materialsandmethods} for further details).

From this generic definition,
we derive two concrete measures of mean polysemy, depending on
how the number of different meanings is calculated for every token:

\begin{APAitemize}
\item \textbf{WordNet polysemy}, this is the number of synsets
(WordNet meanings)
of the word and syntactic category according to the WordNet lexical database.
\item \textbf{SemCor polysemy}, this is the number of
different WordNet synsets associated to a token and syntactic category in the SemCor corpus.
\end{APAitemize}

These two measures of polysemy allow one to capture two extremes:
the full potential number of synsets of a word (WordNet polysemy)
and the actual number of synsets that are used in a corpus (SemCor polysemy),
the latter being a more conservative measure of word polysemy
motivated by the fact that
the number of synsets of a word overestimates, in general,
the number of synsets that are known to an average speaker of English
or the number of synsets to which a child is exposed.

We analyze the dependency between mean polysemy and child age
using data from children involved in longitudinal studies.
Besides the children role, three adult roles were considered as controls:
mothers, fathers and investigators
(see Section \nameref{sec:materialsandmethods} for further details).

In order to check whether the results are a simple effect of the increase of
the production of speech in children as they grow up, we have selected the
first \emph{n} tokens (where $n$ = $50$, $75$, $100$, $125$, $150$, $175$, $250$)
for each conversation.
\textcolor{black}{$n$ is the \emph{sample length}.}
If a \textcolor{black}{transcript} does not contain,
at least, \emph{n} tokens, it is discarded.
Our method of word selection implies that an increase
in the sample length tends to reduce the number of conversations
that can be included in the analysis.
However, a length of $50$ tokens yields results that resemble, qualitatively,
the results that are obtained with other sample lengths.
Thus, we take a length of $50$ tokens as a canonical length.
\textcolor{black}{This implies that all results that
are shown in this article have been obtained with a sample length of $50$ tokens.}

For every individual speaker we compute the mean polysemy
(according to both WordNet and SemCor polysemies)
at each point in time, which is defined
as the age of the individual (in months) in that recording session.
Therefore,  we have a collection of
pairs <time, mean WordNet polysemy> and a collection
of pairs <time, mean SemCor polysemy> for each individual.
We study the evolution of mean polysemy over time from two perspectives:

\begin{itemize}
	\item \textbf{a qualitative analysis}: we average the
	mean polysemy value for all the individuals who have the same role.

	\item \textbf{a quantitative analysis}: for each role, we count the
	number of individuals who show significant (positive or negative)
	or non-significant correlations between mean polysemy and age.
\end{itemize}

The results of the qualitative analysis of the \textbf{WordNet polysemy}
are shown in the left-hand side of Figure
\ref{fig:eng_edat_sinsets_all_50}
and those of \textbf{SemCor polysemy}
are shown in the left-hand side of Figure \ref{fig:eng_edat_sinsets_semcor_all_50}.
In both cases, children exhibit a two-phase (fast-slow) growth of the mean polysemy,
delimited by a breakpoint.
In adults there is no clear positive nor negative tendency.
The breakpoint for mean WordNet polysemy is located at $31.6 \pm 0.1$ months,
and for mean SemCor polysemy, at $31.4 \pm 0.1$ months.
We note that this breakpoint is computed on the average curve for all children
(see Subsection \nameref{subsec:calculationbreakpoint}
for details on how the breakpoint is calculated).

The results of the quantitative analysis of the \textbf{WordNet polysemy}
can be seen in the right-hand side of Figure \ref{fig:eng_edat_sinsets_all_50} and the corresponding
raw data in Table \ref{tab:eng_edat_sinsets_binom_50}
(\textit{All categories} row).
These results confirm those of the qualitative analysis:
almost half of the children ($51.8\%$) show a significant positive correlation
(\emph{$S_+$}),
whereas most adults ($91.1\%$ of the mothers,
$85.7\%$ of fathers and $86.7\%$ of investigators)
show a non-significant tendency.
Notice that the number of significant positive
correlations for children is significantly high ($\uparrow$),
and the number of non-significant correlations is significantly low ($\downarrow$),
both according to a binomial test
(see Subsection \nameref{subsec:binomialtest} for further details).

The results of the quantitative analysis of the \textbf{SemCor polysemy}
can be seen in the right-hand side of Figure \ref{fig:eng_edat_sinsets_semcor_all_50} 
and the corresponding raw data in Table \ref{tab:eng_edat_sinsets_semcor_binom_50}.

In this case, we can see that the trend is the same
as in the case of WordNet polysemy.
We can also see that a remarkably high number of correlations
between mean polysemy and child age are positive and significant in children
($51.8\%$), whereas an overwhelming majority of
these correlations are non-significant in all adult categories
($85.7\%$ for mothers,
$92.9\%$ for fathers and $93.3\%$ for investigators).
Finally, we note that
the average values for mean polysemy
differ depending on the source:
an average polysemy of about $6$ synsets is found in adults
when the source of the polysemy is the SemCor corpus,
compared to an approximate average of about $10$ synsets
when the source is the WordNet database
(see Figure \ref{fig:eng_edat_sinsets_all_50} and
Figure \ref{fig:eng_edat_sinsets_semcor_all_50}).

\textcolor{black}{The analysis of the magnitude of the correlations
provides a complementary perspective, in both WordNet and Semcor polysemy.
The median is about 0.5 in children while it is about zero in adults
(Figure \ref{fig:eng_boxplot_corr}). Correlations are markedly
skewed towards positive values in children. }

\figuraPolysemyAllWordNet
\figuraPolysemyAllSemCor
\figuraBoxPlotAllCorrelations

\figuraPolysemyCategoryWordNet
\figuraPolysemyCategorySemCor
\figuraBoxPlotCategoryCorrelations

\taulaRawDataWordNet
\taulaRawDataSemCor


In order to validate the results that have been computed previously,
in Table \ref{testAOVwordnetX}
we have divided the timeline into ten segments of the same time size.
In each segment we compare the mean polysemy
(both WordNet and SemCor) of children with that of adults.

These results show that the average
WordNet polysemy in children is significantly lower
than that of adults until
the seventh segment (child age between $44.9$ and $50.3$ months)
for WordNet polysemy,
with an F-value that decreases gradually at each segment.
The same pattern can be observed for SemCor polysemy,
but in this case the difference
is significant until the ninth segment (age between $55.8$ and $61.3$ months).
However, as in the previous case, a systematic decrease of the F-value
is observed. In both cases, we need to note that the first segment
has a low F-value, which can be explained by the size of the
sample ($14$ for both WordNet and SemCor polysemy).
These results are coherent with Figures
\ref{fig:eng_edat_sinsets_all_50} and \ref{fig:eng_edat_sinsets_semcor_all_50},
in which it can be seen that WordNet average polysemy in children
is closer to that of adults by the end of the curve with respect to the case of SemCor polysemy.

\taulaTestAnovaWordnetX


We have shown that polysemy tends to increase over 
time in children markedly compared to adults. A bias for low polysemy 
could arise indirectly from a preference for nouns and the fact 
that they have low polysemy compared to verbs
(see \nameref{sec:introduction}).
Therefore, we have controlled for category (part-of-speech) 
to check whether the observed pattern still holds for individual categories. 

We have taken the same amount of tokens from each recording session
but selecting only tokens belonging to one of the four target syntactic categories:
nouns, verbs, adjectives and adverbs.
Tokens that were not in the target category in their context of use,
were discarded.
This implied that, if a session did not contain the required number of
tokens of the same category, it was discarded.

The results of the qualitative analysis of separate syntactic categories
for the mean WordNet polysemy
can be found in the left-hand side of Figure \ref{fig:eng_edat_sinsets_other_50},
the quantitative analyses are shown in
the right-hand side of Figure \ref{fig:eng_edat_sinsets_other_50}
and Table \ref{tab:eng_edat_sinsets_binom_50}.

Comparing these results with those
in Figure \ref{fig:eng_edat_sinsets_all_50},
the tendency for mean polysemy to increase in children seems to blur,
since most children show a non-significant correlation between mean polysemy
and time when single syntactic categories are taken alone.
Only nouns ($8.9\%$) and adverbs ($10.7\%$)
exhibit a relevant number of significant positive correlations (\emph{$S_+$}),
which are in fact significantly high according to a binomial test.
However, these percentages contrast with the percentage of $51.8\%$ 
when all categories were considered together.

A similar tendency is observed when we analyze
the mean SemCor polysemy controlled by category.
The qualitative results
can be found in the left-hand side of
Figure \ref{fig:eng_edat_sinsets_semcor_other_50},
the quantitative results are shown in
the right-hand side of
Figure \ref{fig:eng_edat_sinsets_semcor_other_50}
and Table \ref{tab:eng_edat_sinsets_semcor_binom_50}.
Again, the tendency for mean polysemy to increase in children
is not as clear as when all categories were taken together.
However, in this case, it is found that in all categories,
the percentages of positive significant correlations (\emph{$S_+$})
in children are significantly high according to a binomial test
($8.9\%$ for nouns, $10.7\%$ for verbs, $13\%$ for adjectives
$8.9\%$ and for adverbs),
but, as in the previous case, they are far
from the results in Figure \ref{fig:eng_edat_sinsets_semcor_all_50}
($51.8\%$).

As for adults, the results in both cases show no relevant tendency, 
since in all cases the non-significant correlations form the majority.
This is exactly the same result that was observed
when results were not segmented by syntactic category.

The analysis of magnitudes in Figure \ref{fig:eng_boxplot_corr_cat} 
shows that the skew towards positive correlations in children weakens substantially 
compared to Figure \ref{fig:eng_boxplot_corr}, but magnitudes are still slightly 
biased towards positive values. That was expected to happen for SemCor 
polysemy given Table \ref{tab:eng_edat_sinsets_semcor_binom_50}. 

Therefore, the trend that we observed in the first part of our analyses
dissipates when syntactic categories are considered separately,
but does not disappear completely.
This suggests that there is a preference for words of 
low polysemy actually competing with a preference for nouns. 
To reassure the latter, we have analyzed the evolution of 
the use of part-of-speech categories over time and checked 
that nouns are less polysemous than verbs in our dataset. 

\clearpage

\subsection{Interaction between polysemy and syntactic category}

We have analyzed the evolution of the syntactic categories over time
using the same methodology as in the previous section.
In this case, the variable that is being analyzed
is the percentage of word tokens
of a target category over time.

The results of the qualitative analysis (Figure \ref{fig:eng_edat_category})
show that the percentage of nouns decreases over time in children:
it starts at $80\%$, drops to $40\%$ and then, stabilizes.
Verbs exhibit an opposite tendency in children:
they start at $10\%$ and increase their contribution to $40\%$,
and finally they stabilize.
In fact, both nouns and verbs seem
to stabilize approximately by the same time point.
The breakpoint for the percentage of nouns used is located at $30.0 \pm 0.1$ months
and the breakpoint for the percentage of verbs used is located at $33.0 \pm  0.1$ months
(see Subsection \nameref{subsec:calculationbreakpoint}).

These results are consistent with a well-known phenomenon:
children tend to learn nouns first and then verbs \parencite{Saxton10childlanguage}.
As for the remaining categories (adjectives and adverbs),
Figure \ref{fig:eng_edat_category} suggests
that they remain stable over time.
As looks can be deceiving, stronger conclusions must be
explored with the help of the quantitative analysis.

According to Figure \ref{fig:eng_edat_category_barra},
$41.1\%$ of children show a significant negative correlation
between the mean percentage of nouns and time,
and $66.1\%$ show a significant positive correlation between the mean percentage
of verbs and time, which confirms our observation of the qualitative results.
As for adults, they mostly show non-significant correlations in
all syntactic categories.

\figuraTRES

\figuraQUATRE

Notice that the mean percentage of nouns and verbs stabilize,
in children, by an age ranging between $30$ and $33$ months, that matches
the time interval in which children stabilize their mean WordNet and SemCor polysemy ($31$ months).
This suggests that polysemy and syntactic category could be correlated somehow.

The analysis of the abundance of each category suggest that the overall
tendency of polysemy to increase could be, at least to some extent,
a side-effect of an initial preference
for syntactic categories that have low polysemy 
(see Section \nameref{sec:introduction}).
For this reason, we calculate
the mean polysemy of the tokens of each syntactic category separately,
regardless of time.
We focus on verbs and nouns, as the other two categories that
we have considered (adjectives and adverbs) do not show any relevant tendency,
and they represent only a small percentage (less than $20\%$)
of the total of the yielded tokens.

Table \ref{tab:syn-childes} shows the mean polysemy of the word tokens
that have been produced for each
syntactic category depending on the role of the speaker.
In all roles, verbs have a significantly higher mean polysemy than nouns
(a Fisher randomization test gives a \emph{p}-value $< 10^{-5}$
in all cases; see Subsection \nameref{subsec:fisher}
for further details about the test).

To sum up, our analysis above shows that children exhibit the following behavior:
\begin{APAenumerate}
	
\item Approximately between the 30th and 33rd month, a breakpoint separates two stages:
	  one of quick growth followed by another of gradual convergence to adult linguistic behavior.
	  The breakpoints are located at the $31$st month for WordNet and Semcor polysemies,
	  at the $30$th month for the percentage of nouns, 
	  and at the $33$rd month for the percentage of verbs.

\item Up to the breakpoint, their mean WordNet and SemCor polysemy as well 
	  as their mean percentage of verbs increase while
	  their mean percentage of nouns decreases.

\item From the breakpoint onwards, the mean WordNet and SemCor polysemy as well as
	  the mean percentage of verbs and nouns stabilize and tend to converge on those of adults.
 \end{APAenumerate}

Beyond children, our analysis shows that
\begin{APAenumerate}
\item Verbs have a significantly higher mean polysemy than nouns in all roles.

\item With respect to children, adults show a rather stable 
language production in all analyses over time.

\end{APAenumerate}

In the next section, we explore the implications of
these facts for the origin of the tendency of mean polysemy to increase over time for children.

\taulaDOS

\section{Discussion}

In this article we have investigated the polysemy of words
as a new potential bias of vocabulary learning.
We have shown that the mean polysemy of words
increases over time for children markedly,
but this effect is not observed in adults.
Our findings in children are non-trivial because they differ from the adults 
interacting with them and thus cannot be attributed to Child Directed Speech 
\parencite{Snow1972a,Matychuk2005a}.
The high contrast between the average polysemy of adults and that of young 
children rules out the null hypothesis that children are simply sampling 
tokens uniformly at random from adult speech. Put differently, children 
are of course learning from the speech of the adults in their environment 
but it is not a straightforward imitation process as witnessed by 
the polysemy of the words and other statistical properties of their speech.
However, the gap between child speech and child-directed speech reduces over time, 
suggesting that the null hypothesis may be relevant in the long run. 
The failure of the null hypothesis is consistent with the well-known 
differences between the parameters of Zipf's law in children and 
those of adults \parencite{Piotrowski1995, Baixeries2012c}. 
If children were merely sampling adult speech uniformly at random, 
the parameters would be the same.

Our finding is also unexpected when considered in the light of
principles of language acquisition and facts from quantitative linguistics:
the general bias for high frequency \parencite{Harris2011a} and
{Zipf's law of meaning distribution, i.e.
a positive correlation between frequency and polysemy that concerns
both adult and child language \parencite{Hernandez2016a}.
The polysemy effect cannot be explained by a frequency bias
in a straightforward fashion since it would imply
that children use more polysemous words first,
which would contradict our findings presented here. However,
our findings could be predicted by the law of meaning distribution
and the bias for low frequency that is found across all words
but not within specific lexical categories \parencite{JCL:1917448}.

A critical observation is that the polysemy effect weakens
dramatically (but does not disappear completely)
when controlling for syntactic category
(see Figures \ref{fig:eng_edat_sinsets_other_50}
and \ref{fig:eng_edat_sinsets_semcor_other_50}).
Therefore, we investigated the role that specific
syntactic categories could have in the polysemy effect.
When not focusing on a specific syntactic category,
the effect could be explained through a combination of three facts:

\begin{APAitemize}
\item
	Children decrease their proportion of nouns while
	they increase the proportion of verbs over time
	(recall Figure \ref{fig:eng_edat_category}
	and also \textcite{Saxton10childlanguage}).

\item
	The mean polysemy of verbs is significantly higher than
	that of nouns (recall Table \ref{tab:syn-childes}).

\item
	Nouns and verbs cover the majority of tokens
	that are produced (recall Figure \ref{fig:eng_edat_category}
	and Table \ref{tab:syn-childes}).

\end{APAitemize}

Therefore, we suggest two possible explanations, not necessarily
mutually exclusive,
for the increase in word polysemy over time in children:

\begin{APAenumerate}

\item
	\textbf{Standalone bias}.
	Children have a preference for less polysemous words,
	this is, less ambiguous words. This is supported by Zipf's view of polysemy
	as a cost for the listener \parencite{Zipf1949a}.
	Further support comes from models of Zipf's law that define the listener's
	effort as the entropy of the meanings of a word,
	which can be regarded as a distributional measure of polysemy \parencite{Ferrer2015b}.

\item
	\textbf{Side-effect of other biases}.
	When children learn a language, they begin using more nouns
	than words from other syntactic categories, and then,
	they increase the percentage of verbs
	that they use in their conversations over time.
	Since the mean polysemy of verbs is \textit{significantly}
	higher than that of nouns, the mean polysemy increases because the proportion
	of verbs increases.

\end{APAenumerate}

On the one hand,
the explanation of a side-effect bias is supported
by the fact that the positive correlation between mean polysemy
and child age weakens after controlling for syntactic category.
In particular, $S_+$ (the number of speakers
with a significant positive correlation between mean polysemy and child age)
is only significantly high for nouns in a minority of children
after controlling for syntactic category according to WordNet polysemy
(Table \ref{tab:eng_edat_sinsets_binom_50}).

On the other hand,
the hypothesis of a standalone bias is supported by the
different results seen for SemCor polysemy:
$S_+$ reduces substantially after controlling for syntactic category
but is still significantly high for all syntactic categories
according to SemCor polysemy
(Table \ref{tab:eng_edat_sinsets_semcor_binom_50}).

Therefore, both explanations could be valid to a certain extent.
In fact, an important question for future research is whether one
explanation could be subsumed by the other.
Indeed, the standalone bias could provide
a more parsimonious explanation:
if children prefer less polysemous words in general
they will prefer nouns because they are less polysemous.
However, we may not be able to reduce all preferences
for nouns to polysemy because many nouns are also attractive
for their imageability \parencite{McDonough2011}. 
Besides, we cannot exclude the possibility
that the two explanations above are implications
of a different deeper explanation that has escaped us.

Above we have suggested that the bias for low polysemy could weaken
within specific syntactic categories as a result of it being a side-effect of a preference for nouns.
Another reason for the weakening could be a competition with frequency bias,
that is stronger within specific categories \parencite{JCL:1917448,Hills2010a}
and is in conflict with the low polysemy bias
because the law of meaning distribution predicts that words of low polysemy should
have low frequency.
Finally, another reason could be that the standalone bias applies
to only certain speakers. This raises the broader question of whether
our findings and arguments are valid for all kinds of learners.
Interestingly, learners of a second language show an initial
tendency for WordNet polysemy to increase over time
(this is the conclusion of the primary analysis by \textcite{Crossley2010a}).
This is consistent with our results with L1 learners:
as we have shown, children start learning words that are less polysemous,
according to two different measures of polysemy (WordNet and SemCor).
This suggests that preference for low polysemy
is a bias in the vocabulary acquisition
process that affects both L1 and L2 learners similarly.
An analogous suggestion was made for the bias
by which novel words with many similar sounding
words are learned more quickly \parencite{Stamer2012a}.

The breakpoint by the age of 30-33 months in the evolution of polysemy 
and the percentage of syntactic categories
in children coincides in the timeline with the end of the critical
period, by the same age, which is traditionally assigned to syntactic development
\parencite{Kuhl2010, Yule2006, Pinker1994},
and the production of closed-class words \parencite{JCL:1917448}.
We believe that the relationship between the stabilization of
mean polysemy of children and milestones in the evolution of
child language should be the subject of future research.

\section{Conclusions}

We have studied the evolution of the polysemy in children,
and have put forward polysemy as a learning bias in their vocabulary acquisition.
Our main conclusions are:

\begin{enumerate}

\item
	There is a non-trival pattern in the evolution of polysemy over time.
	Children increase their mean polysemy
	in two phases:
	an initial phase with a fast growth of polysemy and a second phase
	with a slower growth of polysemy.
	In contrast, adults interacting with them do not show this tendency.

\item
	This non trivial pattern weakens when the analysis
	is segmented by syntactic category.

\item
	Children show a tendency to learn nouns first
	and then verbs, which is consistent with previous research
	\parencite{JCL:1917448, Gentner2006, Gentner1982}.

\item
	Verbs have a significantly higher mean polysemy than nouns
	in all roles: children and adults.

\item
	The last two facts could explain the pattern of
	the evolution of polysemy over time to some extent.

\item
	That role of a standalone bias for low polysemy cannot be discounted.

\item
	Our findings and \textcite{Crossley2010a} suggest that L1 and L2 learners
	resemble each other in their dominant biases on polysemy:
	a preference for low polysemy words prevails in both kinds of learners.
	
\end{enumerate}

A deeper understanding of why the bias for low polysemy weakens within
specific syntactic categories and how it interacts with frequency is a challenge for future research.
Also, the relationship between the senses that
a speaker really knows about a word and its potential number of synsets
should be investigated in detail.

\section*{Acknowledgements}

This research work was acknowledged by the 
2017SGR-856 (MACDA) project from AGAUR (Generalitat de Catalunya),
and supported by the grant TIN2017-89244-R 
from MINECO (Ministerio de Econom{\'i}a y Competitividad).

\ifinteract
\else
\bibliography{biblio.bib}
\bibliographystyle{plain}
\fi

\end{document}